# Intuitions and the Modelling of Defeasible Reasoning: some Case Studies

**Henry Prakken**
Institute of Information and Computing Sciences
Utrecht University
Utrecht, The Netherlands
`henry@cs.uu.nl`
`http://www.cs.uu.nl/staff/henry.html`

## Abstract

The purpose of this paper is to address some criticisms recently raised by John Horty in two articles against the validity of two commonly accepted defeasible reasoning patterns, viz. reinstatement and floating conclusions. I shall argue that Horty's counterexamples, although they significantly raise our understanding of these reasoning patterns, do not show their invalidity. Some of them reflect patterns which, if made explicit in the formalisation, avoid the unwanted inference without having to give up the criticised inference principles. Other examples seem to involve hidden assumptions about the specific problem which, if made explicit, are nothing but extra information that defeat the defeasible inference. These considerations will be put in a wider perspective by reflecting on the nature of defeasible reasoning principles as principles of justified acceptance rather than 'real' logical inference.

## 1 Introduction

The purpose of this paper is to address some criticisms recently raised by John Horty in two articles (Horty; 2001, 2002) against the validity of some commonly accepted defeasible reasoning patterns. Horty criticises two such patterns, viz. reinstatement and floating conclusions. In addition, he criticises a group of inferences where the failure to interleave argument construction and argument evaluation would lead to problems. For each of these cases, Horty presents a series of examples in which his intuitions are that it is coherent for a reasoner to accept the premises but not the conclusion.

Horty's examples are intriguing and his discussion of them is very insightful. Yet I disagree with the conclusions he draws from them. I shall argue that his examples do not demonstrate the invalidity of the reasoning patterns. Some of them reflect patterns which, if made explicit in the formalisation, avoid the unwanted inference without having to give up the inference principles criticised by Horty. In the remaining examples Horty's intuitions seem to be based on hidden assumptions about the specific problem which, if made explicit, are nothing but extra information under which the inference cannot be drawn; in other words, these examples do not show that the reasoning patterns are invalid but just that they are defeasible. I shall then put these considerations in a wider perspective, reflecting on the nature of defeasible reasoning principles as principles of justified acceptance rather than 'real' logical inference.

The discussion in this paper will be informal and conceptual rather than technical, and will freely switch from one style of nonmonotonic logic to another whenever appropriate (although with some bias to argumentation approaches). I assume that the reader is familiar with the essentials of the main approaches to nonmonotonic logic, especially default logic, preferential entailment and argumentation systems (particularly in the style of Pollock (1995) and Dung (1995)). Formal treatments of most examples of this paper can be found in the two papers by Horty and in (Prakken and Vreeswijk; 2002).

First I will say more on the role of intuitions in the development of logic in general and of nonmonotonic logic in particular.

## 2 Intuitions in logic

In both his articles Horty mainly relies on intuitions in concrete examples. In fact, he explicitly says that this is "the only method that I know of in this area, where there is no recourse to anything like a formal semantics: (...)" (Horty; 2001, p. 9). In fact, in the field of nonmonotonic logic this approach has been very popular; see e.g. Lifschitz (1988)'s well-known list of benchmark examples. Setting aside the issue whether a formal semantics of defeasible reasoning is indeed impossible, I will argue that in this area relying on intuitions is extremely prone to error, and that there are better, more principled approaches.

The use of intuitions in evaluating logical systems has been criticed before; see e.g. Veltman (1985). For one thing, the question arises whose intuitions should count. Those of logicians are hopelessly corrupted by overexposure to formalism. So should we ask the 'average language user', hoping that they are not infected by theoretical bias? Then the problem often arises that their answers reveal a lacking understanding of the reasoning patterns; however, teaching them about these reasoning patterns infects them with the theoretical bias we were hoping to avoid. Another obvious problem is that intuitions of different persons often conflict. For these and other reasons a sensible way to use intuitions seems, each time they present us with a puzzle, not to directly rely on them, but to see whether they reflect some underlying pattern. This is the strategy I will follow in the present paper.

## 3 Reinstatement

The first reasoning pattern discussed by Horty (2001) is reinstatement. This pattern, endorsed by most nonmonotonic logics, can be naturally explained in terms of argumentation systems.

Let me first fix some terminology on argumentation systems that I will use throughout the paper, sometimes applied to other nonmonotonic logis as well (such application is possible since, as shown by e.g. Dung (1995) and Bondarenko et al. (1997), many nonmonotonic logics can be reformulated as argumentation systems). *Defeat* is a binary relation between arguments: that argument $A$ defeats argument $B$ means that $A$ and $B$ are in conflict with each other and $B$ is not preferred over $A$. I say that $A$ *strictly defeats* $B$ if $A$ defeats $B$ and not vice versa.

*Justification* is a skeptical notion of nonmonotonic consequence. Argumentation systems usually classify argments into three classes, justified, defensible and overruled arguments. Loosely speaking, the *justified* arguments are those with which a dispute can be 'won', the *overruled* arguments are those with which a dispute should be 'lost', and the *defensible* arguments should leave the dispute undecided. Conclusions of justified arguments are skeptical consequences and those of defensible arguments are credulous consequences of a theory. Horty only discusses skeptical consequence, i.e., justification.

Now *reinstatement* is the phenomenon that an argument that is (perhaps strictly) defeated by another argument is still justified since it its defeater is itself strictly defeated by another justified argument; cf. Figure 1.

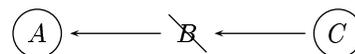

Figure 1: Argument $C$ reinstates argument $A$.

Reinstatement is embodied in all Dung-style argumentation systems via a notion of *acceptability*: an argument $A$ is acceptable with respect to a set $S$ of arguments if all arguments defeating $A$ are defeated by an argument in $S$. For instance, in Figure 1 the argument $A$ is acceptable with respect to the set $S = \{A, C\}$.

Reinstatement has a *direct* and an *indirect* form, and Horty only rejects its direct form. Direct reinstatement is when all three arguments are in conflict on their final conclusions, as in the following example (where the more specific default is preferred over the less specific one).

**Example 3.1**

$A$: Tweety flies because it is a bird
$B$: Tweety does not fly because it is a penguin
$C$: Tweety flies because it is a magic penguin

Indirect reinstatement, on the other hand, is when the reinstating argument $C$ defeats the 'middle' argument $B$ on one of its intermediary conclusions, such as in the following example.

**Example 3.2**

$A$: Tweety flies because it is a bird
$B$: Tweety is a penguin because it was observed to be so, so Tweety does not fly
$C$: The observation that Tweety is a penguin is unreliable since it was done during a blizzard

Figure 2 depicts indirect reinstatement (as for notation, for any pair of arguments $X$ and $X^-$, the latter is a proper subargument of the first.)

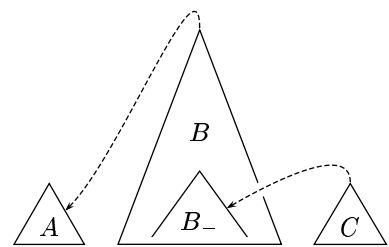

Figure 2: Indirect reinstatement

### 3.1 Is direct reinstatement invalid?

Horty (2001) argues that Example 3.1 shows that direct reinstatement is invalid, since the reason that Tweety flies is that it is a magic penguin and not that it is a bird. (He does so in a critical discussion of (Prakken and Sartor; 1997), a system of prioritised extended logic programming that instantiates Dung (1995)'s grounded semantics.)

Before Horty reaches this conclusion, he first considers an alternative strategy with such examples, viz. to simply live with the anomaly that the first default is not defeated on the grounds that after all the correct conclusion is drawn that Tweety can fly. He rejects this strategy with an example in which the consequent of the least preferred default is logically stronger than that of the most preferred default.

**Example 3.3**

- $d_1$: Microsoft employees tend to be millionaire
- $d_2$: New Microsoft employees tend to have less than half a million
- $d_3$: New Microsoft employees in department $X$ have at least half a million.

Suppose again that the more specific default is preferred over the less specific one. According to Horty, in this case an argument for new Microsoft employees in department $X$ based on the least specific default $d_1$ leads to the incorrect conclusion that they are millionaires. Horty therefore concludes that direct reinstatement must be rejected as a principle of defeasible reasoning.

How convincing is Horty's argument? I agree with him that in Examples 3.1 and 3.3 the least specific default should not be applied, so Horty is right when he says that reinstatement cannot be combined with the simple representation method of Example 3.3. In this sense I agree with his criticism on my earlier work in (Prakken and Sartor; 1997). However, I think that these observations do not imply invalidity of direct reinstatement, since there is an alternative approach which respects Horty's observations in the examples but validates reinstatement, and which can make relevant distinctions between kinds of examples that go unnoticed in the simple representation scheme of Example 3.3.

The alternative approach is to make the language expressive enough to let applicable defaults block the applicability of other defaults when needed. One way to implement this approach is well-known technique of exception or abnormality clauses (note that the logic of (Prakken and Sartor; 1997) supports this technique, although in most of our examples we did not use it). For instance, in preferential entailment with minimisation of the $Ab$ predicate, all preferred models of the following theory plus the fact that Jeff is a new Microsoft employee of department $X$ satisfy that Jeff owns at least half a million dollars, but not all such models satisfy that he is a millionaire.

- $r_1$: *Microsoft employee* $\wedge \neg Ab_1 \to\ \geq 1M$
- $r_2$: *New Microsoft Employee* $\to$ *Microsoft employee*
- $r_3$: *New Microsoft employee* $\wedge \neg Ab_3 \to\ < 1/2M$
- $r_4$: *New Microsoft employee* $\to\ Ab_1$
- $r_5$: *New MicrosoftX Employee* $\to$ *New Microsoft employee*
- $r_6$: *New MicrosoftX employee* $\wedge \neg Ab_6 \to\ \geq 1/2M$
- $r_7$: *New MicrosoftX employee* $\to\ Ab_3$

In argumentation terms, the argument for $Ab_1$ is not affected by the argument for $Ab_3$: a new Microsoft employee is still an abnormal Microsoft employee, but he may also be an abnormal new Microsoft employee.

The reason that this approach seems better than simply invalidating reinstatement is that whether reinstatement should go through or not seems to depend on the nature of the domain, the kind of knowledge involved and the context in which this knowledge is used. Consider the following moral example on reasons for the severity of punishment.

**Example 3.4**

- $r_1$: For theft imprisonment upto 6 years is acceptable
- $r_2$: For theft out of poverty imprisonment of more than 3 years is not acceptable
- $r_3$: For theft during riots only imprisonment of more than 4 years is acceptable

And suppose that $r_3$ is preferred to $r_2$ which is in turn preferred to $r_1$. As in the Microsoft example, the consequent of the most preferred default is logically weaker than the consequent of the least preferred default. Yet it seems that for theft during riots out of poverty any imprisonment must be between 4 and 6 years. A general reason for this outcome could be that reasons for the severity of punishment do not block but just outweigh each other.

Or consider the following legal evidence example.

**Example 3.5** Consider the following three witness statements.

- $w_1$: John says that the suspect stabbed the victim
- $w_2$: Bob says that the suspect did not kill the victim
- $w_3$: Al says that the suspect killed the victim

And suppose that for whatever reason Bob is regarded as more credible than John and Al as more credible than Bob (suppose, for instance, that Al is always right). Here it seems quite reasonable to accept the least credible source since the only defeating source is overruled.

The point of these examples is that in modelling default reasoning one needs the extra expressiveness in order to control the blocking of defaults. Of course, one should avoid tinkering with the formalisation in concrete examples just to get a desired outcome. Therefore, the challenge for research is to find general principles for choosing the right formalisation. For two examples of research in this spirit see Pollock's research on undercutting defeaters in epistemic contexts, e.g. Pollock (1995), and Hage and Verheij's work on defeasible reasoning with legal rules and principles, e.g. Verheij et al. (1998). This work indicates that such principles depend on the domain, the kind of knowledge involved and the context in which it is used. For instance, in statistical reasoning with empirical regularities blocking more general defaults by more specific defaults can be justified by the total evidence requirement of statistical reasoning. Now a crucial difference between the three examples of this section is that only Example 3.3 involves reasoning of the latter kind; Example 3.4 involves moral reasoning, while Example 3.5 involves reasoning about the credibility of information sources.

Concluding this section, I agree with Horty that direct reinstatement does not go together with the simple representation scheme of Example 3.3 (used by Horty (1994) in his inheritance systems and by me and Giovanni Sartor in most of (Prakken and Sartor; 1997)). However, I have argued that the problems are not due to direct reinstatement but to the weak expressiveness of the representation. This conclusion can be strengthened if further defeasible reasoning patterns can be found where more expressiveness of the above kind is needed. And below I will argue that such further reasoning patterns indeed exist.

### 3.2 Should argument construction and argument evaluation be interleaved?

Horty (2001) discusses a further set of examples which according to him reveals another flaw of Dung-style systems, viz. that they don't interleave the construction and evaluation of arguments. However, I will argue that these examples are better treated as attacks on reinstatement.

**Example 3.6** Consider four arguments $A, B, C$ and $D$ such that $B$ strictly defeats $A$, $D$ strictly defeats $C$, $A$ and $D$ defeat each other and $B$ and $C$ defeat each other.

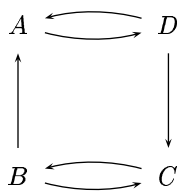

Horty discusses the following instantiation, again with subclass-superclass defeat.

$A$: Larry is rich because he is a public defender, public defenders are lawyers, and lawyers are rich
$B$: Larry is not rich because he is a public defender, and public defenders are not rich
$C$: Larry is rich because he lives in Brentwood, and people who live in Brentwood are rich
$D$: Larry is not rich because he rents in Brentwood, and people who rent in Brentwood are not rich

Since none of these arguments is undefeated, in none of the Dung-style systems it skeptically follows that Larry is poor. The same holds for prioritised default logics (e.g. Brewka (1994)). Yet it might be argued that since both arguments that Larry is rich are strictly defeated by an argument that Larry is not rich, it should skeptically follow that Larry is not rich. This is the outcome obtained by Horty (1994).

Horty's diagnosis is that the problems are caused by mistakenly separating the phases of argument construction and argument evaluation. According to him, Dung-style systems reflect a picture of defeasible reasoning in which first all possible arguments are constructed, all relevant defeat relations are identified, and then the justified arguments are determined on the basis of this overall pattern of defeat. Horty prefers an approach in which "arguments are constructed step-by-step and are evaluated in each step of their construction: those that are indefensible (...) are discarded at once, and so cannot influence the status of others." (Horty; 2001, p. 11). Applied to the example, in this approach the arguments $A$ and $C$ are discarded as soon as their defeat by $B$, respectively, $D$ is noted, so that they cannot interfere with $D$, respectively, $B$.

To comment on this, note first that the interleaving approach is consistent with the validity of direct reinstatement. Suppose a reasoner, when confronted with the arguments $A$ and $B$, aks whether $A$ is now "clearly indefensible". The answer is "yes" only if the fact that $B$ is in turn defeated by $C$ is regarded irrelevant for the status of $A$, so only if reinstatement is rejected; but the interleaving approach does not force this answer. So, Horty's examples seem more convincing as examples against reinstatement than against 'non-interleaving'.

Now does the example reveal a flaw with direct reinstatement? The same alternative approach as above prevents a clear positive answer:

$r_1$: *Lawyer* $\wedge \neg Ab_1 \to$ *Rich*
$r_2$: *Public defender* $\to$ *Lawyer*
$r_3$: *Public defender* $\wedge \neg Ab_3 \to \neg$ *Rich*
$r_4$: *Public defender* $\to$ $Ab_1$
$r_5$: *Brentwood resident* $\wedge \neg Ab_5 \to$ *Rich*
$r_6$: *Brentwood tenant* $\to$ *Brentwood resident*
$r_7$: *Brentwood tenant* $\wedge \neg Ab_7 \to \neg$ *Rich*
$r_8$: *Brentwood tenant* $\to$ $Ab_5$

The point is that $r_4$ and $r_8$ give rise to two additional arguments $E$ and $F$, undercutting, respectively, $A$ and $C$, so that $B$ and $D$ are justified. Again the undercutters are based on the general principle that statistical information about subclasses overrides conflicting statistical information about superclasses.

### 3.3 Conclusion on reinstatement

What conclusions can be drawn fom the discussion so far? I have said that there are good reasons not to reject direct reinstatement but instead to adopt a logic of sufficient expressiveness in which reinstatement is valid but can be avoided. With the latter I mean that the logic should offer the means to formalise, when desirable, reinstatement-like examples in a way that does not formally reflect the reinstatement pattern. This approach should be combined with an investigation of general principles for choosing the proper formalisation.

## 4 Floating conclusions

I now turn to Horty (2002)'s criticism of another defeasible reasoning pattern. Sometimes one is faced with two conflicting and equally strong arguments, but whatever way the conflict would be resolved, a certain conclusion could always be reached. Such conclusions were by Makinson and Schlechta (1991) called 'floating conclusions'. When modelling skeptical reasoning, the question is whether floating conclusions should come out as justified.

Floating conclusions can only be modelled as justified if the logic gives rise to alternative conclusion sets in case of an irresolvable conflict (such as default logic or Dung's preferred and stable argumentation semantics). In such logics the justified conclusions can be defined as those contained in all such sets. If a logic always induces a unique conclusion set (such as Dung (1995)'s grounded semantics and Horty (1994)'s inheritance system), then floating conclusions cannot be recognised as justified. This is generally regarded as a drawback of such systems.

Let us consider an example.

**Example 4.1** (Floating conclusions.) Consider the arguments $A^-$, $A$, $B^-$ and $B$ such that $A^-$ and $B^-$ defeat each other and $A$ and $B$ have the same conclusion.

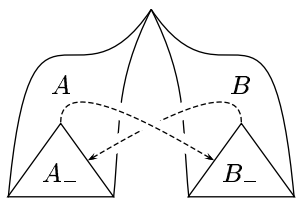

An intuitive reading (entirely based on reality) is

$A^-$: Brygt Rykkje is Dutch since he was born in Holland
$B^-$: Brygt Rykkje is Norwegian since he has a Norwegian name
$A$: Brygt Rykkje likes ice skating since he is Dutch
$B$: Brygt Rykkje likes ice skating since he is Norwegian

Whichever way the conflict between $A^-$ and $B^-$ is decided, we always end up with an argument for the conclusion that Brygt Rykkje likes ice skating, so it seems that this conclusion is justified, even though it is not supported by a justified argument. In other words, the status of this conclusion floats on the status of the arguments $A^-$ and $B^-$.

It is easy to see that Horty's 'interleaving' approach to defeasible reasoning fails to draw this floating conclusion, since it cuts off both arguments after $A^-$ and $B^-$. However, it is important to note that rejecting floating conclusions does not imply a rejection of direct reinstatement. There are natural systems (such as grounded semantics for Dung's argumentation systems) that validate reinstatement but that do not validate floating conclusions.

### 4.1 Horty examples (1): counterexamples or additional information?

In the above example, there seems nothing wrong with accepting the floating conclusion as justified. However, Horty (2002) presents an intriguing set of other instantiations, in all of which the situation is less clear. Horty therefore concludes that it is at least coherent for a skeptical reasoner not to accept floating conclusions, so that such conclusions cannot be accepted as a matter of logic.

Horty's main example is a rather dramatic one, in which a person (call him Bob) must quickly decide about a certain investment, and whose parents are both terminally ill. Bob knows that both his parents are rich, but he does not know whether they will let him inherit a large sum of money. His sister tells him that she has spoken to both her parents, and that father will not let him inherit anything but mother will let him inherit enough to make his investment. His brother tells him the reverse, viz. that he has also spoken to both his parents, and that not his mother but his father will let him inherit enough to make his investment. Horty argues that in this example it is reasonable to withhold any judgement on whether the investment is secure, since brother and sister's conflicting testimonies undermine each other's credibility; and why should it be more reasonable to believe one of them than to believe neither?

The examples highlight a very interesting pattern, viz. the pattern where the fact that sources of information con-flict undermines the credibility of both sources. However, I think that this specific example has some weaknesses, which indicate the problematic nature of intuitions in concrete examples. The point is that it is very easy to read an additional default principle into the example, viz. that people tend to speak the truth about their intentions, and that it is also very easy to mistake this additional information for the intuition that the reasoning pattern is invalid. Note that this additional default principle is undercut as soon as it turns out that a person has told conflicting things about his or intentions to different persons. Now if both the additional principle and its undercutter are made explicit, there will also be conclusion sets where neither father nor mother lets Bob inherit anything, so that the issue of floating conclusions does not arise.

The same analysis applies to several other of Horty (2002)'s examples. For instance, Horty discusses an example with two defaults saying that persons tend to live where their spouse lives and that they tend to live where they work. He applies this to a case where a person works in X while his wife lives in Y, and he observes that the conclusion that the person lives in X or Y is less likely than the conclusion that he lives somewhere in between X and Y. Again I agree, but I think that Horty fails to make another default explicit, viz. that couples who both work but not in the same city, tend to live somewhere in between where they work. If this default is added, the unwarranted disjunctive conclusion is not justified any more.

The main lesson to be learned from these examples is that in modelling defeasible reasoning it is very dangerous to rely on intuitions in concrete examples. In fact, counterintuitions are much more problematic in defeasible than in deductive reasoning. In the latter, any counterexample invalidates an inference, precisely because the inferences are meant to be deductive. But when extra information can invalidate an inference, the boundary between being a counterexample and being additional invalidating information becomes vague.

However, there are also examples to which the above criticism does not seem to apply with the same force.

### 4.2 Horty examples (2): mutually undermining floating conclusions

Horty also discusses examples where it is less easy to point at suppressed additional information. In one example, two military spies report to their commander on their enemy: one spy says that the enemy has retreated to the forest, the other that the enemy has retreated to the mountains. Should the commander believe that the enemy has retreated and act accordingly? In this example I agree with Horty that the fact that the two spy reports conflict undermines the credibility of both, so that the commander is not justified in believing that the enemy has retreated.

For a legal version of this example, consider the following two witness testimonies.

**Example 4.2**

$w_1$: John says that the suspect stabbed the victim
$w_2$: Bob says that the suspect shot the victim

Clearly, the two witnesses undermine each other's credibility, so in the absence of further information on their credibility the conclusion that the suspect killed the victim is not warranted.

However, again I disagree that these examples simply show the invalidity of a reasoning pattern. In fact, the same strategy works as when reinstatement seems undesirable: whenever floating conclusions appear unacceptable, formalise the example in such a way that it does not instantiate the pattern of floating conclusions.

Of course, this strategy makes sense only if the best formalisation can be chosen on the basis of general criteria. Now in all of these examples the key issue is reliability of a certain source of information or advice. So we should look for general criteria for how the credibility of witnesses or experts can be undercut. One natural principle is that the fact that two witnesses or experts contradict each other undercuts the credibility of both. In Pollock (1995)'s system, which formalises the interaction between defeasible reasons and their possible undercutters, this could be expressed as follows.

$r_1$: If witness $w$ says $p$, this is a prima facie reason for believing $p$
$r_2$: if witness $w'$ says $q$ and $p$ and $q$ are incompatible, this is a prima facie reason for believing $\neg r_1$

$r_2$ is a (defeasible) undercutter of $r_1$. It is easy to see that in case of John and Bob $r_2$ applies to both of them so $r_1$ is blocked for of them.

However, in Section 3 I argued that a third witness statement might reinstate one of Bob or John's testimony. Consider again Example 3.5 and assume that for certain particular reasons Al's statement is judged more credible than Bob's:

$w_1$: John says that the suspect stabbed the victim
$w_2$: Bob says that the suspect did not kill the victim
$w_3$: Al says that the suspect killed the victim
$r_3$: $w_3$ and *special reasons* are a prima facie reason for believing $\neg r_2$ as applied to Bob.

$r_3$ is an undercutter of the instance of $r_2$ that applies to Bob. The result in Pollock's system is that not only the argument

based on Al's testimony but also the argument based on John's testimony is justified, so that we are warranted in believing not only that the suspect killed the victim but also that he shot him.

So far I have put a lot of effort in avoiding floating conclusions by formalising examples in a certain way. Of course, this effort is worthwhile only if there are also also examples where floating conclusions should be accepted, otherwise I might just as well adopt a logic that simply invalidates them. Example 4.1 is such an example. The defaults in that example are just statistical regularities; the fact that they conflict does not undermine the fairness of the samples on which they are based, so they do not undercut each other. Another example where conflicting defaults do not undermine each other is conflicting interpretations of a legal norm. If both interpretations lead, for instance, to the conclusion that someone has committed a tort, there seems no reason for the judge to choose.

### 4.3 Concluding

Concluding, I have shown that skeptical reasoning in multiple-extension logics with a sufficiently expressive language can distinguish between cases where floating conclusions appear to be justified and cases where they appear not to be justified. Of course, it is debatable whether the patterns identified by me are indeed relevant. Therefore, I will next discuss how Horty's approach fares if the pattern of floating conclusions is rejected.

## 5 Zombie arguments

Let us for the sake of argument suppose that floating conclusions should not be accepted as justified. Is Horty's approach then unproblematic, or are there other reasoning patterns where it runs into problems?

Following Makinson and Schlechta (1991) I think there is such a pattern, viz. so-called 'zombie arguments'. To explain this, Horty's approach must be described in more detail. Recall from Section 3 that Horty regards defeasible reasoning as bottom-up construction of arguments interleaved with argument comparison: moreover, arguments are compared in a "deeply skeptical" way: as soon as a counterargument is found that is at least not weaker, an argument is cut off. In case two counterarguments are equally strong, this means that both arguments are cut off. This explains why Horty's approach invalidates floating conclusions, since in the relevant examples construction of an argument is cut off before it reaches the floating conclusion.

Now this "deeply skeptical" approach deals in a peculiar way with another type of example.

**Example 5.1** (Zombie arguments.) Consider the arguments $A^-$, $A$, $B$ and $C$ such that $A^-$ and $B$ defeat each other and $A$ defeats $C$.

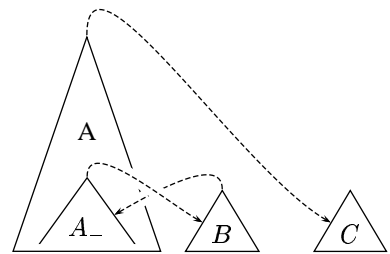

A concrete example is

- $A^-$: Dixon is a pacifist since he is a quaker
- $B$: Dixon is no pacifist since he is a republican
- $A$: $A^-$ + Dixon has no gun since he is a pacifist
- $C$: Dixon has a gun since he lives in Chicago

Suppose further that the first two defaults in the example are equally strong so that $A^-$ and $B$ defeat each other, while the third default is stronger than the fourth so that $A$ strictly defeats $C$. In Horty's approach we must conclude that Dixon has a gun, since the only potential counterargument, $A$, is cut off after detecting that its subargument $A^-$ is defeated by $B$.

Here Horty's deep skepticism seems self-defeating, since it makes an argument justified that is not justified in an approach that Horty regards as a weaker form of skepticism, viz. to consider all ways to resolve the conflict between $A^-$ and $B$. Makinson and Schlechta (1991) call arguments like $B$ 'zombie' arguments: $B$ is not 'alive', (i.e., not justified) but it is not fully dead either; it has an intermediate status (defensible), in which it can still influence the status of other arguments.

Now is someone who does not want to accept floating conclusions forced to accept problems with zombie paths? This is not the case, since there are forms of skepticism that reject floating conclusions but recognise zombie arguments. For instance, in Dung's preferred or stable semantics one could define the justified conclusions to be those of arguments that are in all extensions. Or consider Dung's grounded semantics, which regards as justified all arguments in the least fixpoint of an operator that for any set of arguments returns all arguments that are acceptable to it. In the example this operator, when applied to the empty set, returns the empty set since none of the arguments is undefeated. Now, interestingly, these 'moderate' forms of skepticism validate direct reinstatement, so that it seems hard to reject direct reinstatement without incurring problems with Zombie arguments.

Finally, how does the alternative approach deal with Zom-bie arguments? Recall that this approach consists of accepting both direct reinstatement and floating conclusions while providing the means to formalise examples such that they are not instances of these patterns. Now interestingly, in cases where I agree with Horty that floating conclusions should be avoided, zombie arguments might remain undetected. To see this, extend the military example with two defaults, viz. that in general the mountains are safe, but that if the enemy is in the mountains, the mountains are not safe. Then we have the following three arguments:

**Example 5.2**

- $A^-$: The enemy has retreated to the mountains since spy Al says so
- $B$: The enemy has retreated to the forest since spy Bob says so
- $A$: $A^-$ + The mountains are not safe since the enemy is in the mountains
- $C$: By default, the mountains are safe.

If the two spy defaults are blocked as above, then the zombie argument $A$ is not constructible, so $C$ comes out as justified.

Is this a problem for the alternative approach? I don't think so, because of the same difference in underlying pattern as we have seen above: while Example 5.1 involves empirical regularities, Example 5.2 involves credibility of information sources. And in the latter case it seems reasonable to accept that two conflicting spy reports undermine each other if there is no reason to prefer one of them.

## 6 Conclusion

In this paper I have examined whether Horty's counterexamples against reinstatement and floating conclusions have provided convincing reasons to reject the validity of these reasoning patterns. In the course of my discussion, I have also examined the method by which Horty evaluates nonmonotonic logics, viz. assessing them in the light of intuitions concerning concrete examples. I now summarise the main conclusions drawn in this paper, and then put them in a wider perspective.

### 6.1 On reinstatement and floating conclusions

To start with, we can say that Horty's discussion of reinstatement and floating conclusions has significantly raised our understanding of these reasoning patterns: Horty has rightly pointed at problems with naively accepting them. However, I have argued that his examples do not yield convincing reaosns to reject these patterns. A better conclusion from Horty's examples is that nonmonotonic logics should provide the expressiveness to make relevant distinctions between classes of examples. More precisely, I have argued that nonmonotonic logics should validate at least reinstatement and perhaps also floating conclusions, but provide the means to formalise examples such that they do not instantiate these patterns. This approach should be complemented with investigating general criteria for choosing the proper formalisation.

An additional reason for this approach is that a complete rejection of reinstatement leads to a failure to recognise so-called zombie arguments, which failure seems to indicate that Horty's deeply skeptical approach is in fact self-defeating. On the other hand, we have seen that rejecting floating conclusions does not imply a rejection of reinstatement so that Horty's arguments against the former pattern do not reinforce his reasons for rejecting the latter pattern.

### 6.2 On Horty's method of evaluating nonmonotonic logics

As I said above, Horty claims that the only suitable way to evaluate nonmonotonic logics is applying them to concrete examples and checking whether they validate one's intuitions concerning these examples. In Section 2 I already hinted at some general drawbacks of this approach, and I have illustrated these drawbacks in this paper with several examples in which intuitions seemed to be confused with implicit information which, when made explicit, makes the counterintuitiveness disappear. This confusion can very easily arise in defeasible reasoning, since a hallmark of such reasoning is that inferences that are valid relative to certain information can be invalidated by further information. In my opinion it is therefore better to use intuitions not as critical tests but as generators for further investigation. One should always first ask whether perhaps the counterexample is based on suppressed additional information. If that is not the case, one should ask whether the counterexample reflects a general pattern that can be made explicit by enriching the language.

### 6.3 On the nature of principles of defeasible reasoning

I end my conclusions with some speculative remarks on the nature of defeasible reasoning principles, which are meant to put the foregoing in a wider perspective. These remarks are partly repeated from (Prakken and Vreeswijk; 2002), in some agreement with e.g. Loui (1998) and Schlechta (2000) and perhaps going back to Toulmin (1958). Interestingly, towards the end of his (2002) paper Horty comes very close to endorsing a similar position.

For the most part, Horty's two papers reflect a traditional picture of nonmonotonic logic as essentially just another kind of logic, defining validity of inferences, so that the question whether a certain reasoning pattern should be endorsed solely depends on the meaning of the logical operators involved. This explains his attitude that one counterexample suffices to reject the validity of a defeasible reasoning pattern. However, a different view is possible, viz. to simply accept that defeasible reasoning is logically invalid reasoning: after all, nonmonotonic conclusions of a theory are not satisfied by all models of the theory. This leads to a different, more pragmatic picture of the status of defeasible reasoning patterns as principles of justified acceptance, i.e., criteria for when a statement can be accepted on the basis of a theory given that it does not logically follow from the theory. Put differently, since defeasible reasoning is "quick and dirty" reasoning (Schlechta; 2000), we cannot expect a neat and clean theory of such reasoning; the best we can hope for is heuristics, and heuristics are not invalidated by single counterexamples.

If this perspective is taken, room is made for other than just syntactic and semantic criteria: what is also important is, for instance, the nature of the domain and the knowledge employed (e.g. statistical, causal, moral, legal) and the context in which this knowledge is used (e.g. scientific reasoning, quick commonsense planning, a law suit, a medical treatment). This paper's approach with exception clauses provides the means to express at least some such considerations in the object level language, so that their 'nonlogical' nature is respected. This perspective can also do justice to the dynamic aspects of defeasible reasoning (cf. Schlechta (2000)); for instance, a judge faced with two conflicting witnesses as in Example 4.2, would not yet draw a conclusion but instead ask the witnesses "did you hear any sound?".